\documentclass[10pt,twocolumn,letterpaper]{article}

\usepackage[pagenumbers]{cvpr} 

\definecolor{cvprblue}{rgb}{0.21,0.49,0.74}
\usepackage[pagebackref,breaklinks,colorlinks,allcolors=cvprblue]{hyperref}

\usepackage{float}
\usepackage{graphicx}
\usepackage{booktabs}
\usepackage{multirow}
\usepackage{array}

\usepackage{algorithm}
\usepackage{algpseudocode}
\newcolumntype{C}{>{\centering\arraybackslash}p{2.2em}}

\usepackage[most]{tcolorbox}
\usepackage{multicol}
\usepackage{cuted}

\tcbset{
  promptbox/.style={
    colback=gray!5,
    colframe=black,
    boxrule=0.5pt,
    arc=2pt,
    left=6pt,right=6pt,top=6pt,bottom=6pt
  }
}

\def\paperID{} 
\def\confName{CVPR}
\def\confYear{2026}

\title{VideoMemory: Toward Consistent Video Generation via Memory Integration}

\def\authorBlock{
    Jinsong Zhou\textsuperscript{$1,3*$} \qquad
    Yihua Du\textsuperscript{$1*$}  \qquad
    Xinli Xu\textsuperscript{$1*\dagger$} \qquad
    Luozhou Wang\textsuperscript{$1$} \quad
    Zijie Zhuang\textsuperscript{$1$} \quad \\
    Yehang Zhang\textsuperscript{$1$} \quad
    Shuaibo Li\textsuperscript{$1$} \quad
    Xiaojun Hu\textsuperscript{$3$} \quad
    Bolan Su\textsuperscript{$3$} \quad
    Ying-cong Chen\textsuperscript{$1,2\ddagger$}\quad
    
    \\
    \small$^1$ HKUST(GZ)\quad 
    \small$^2$ HKUST\quad 
    \small$^3$ ByteDance\quad

    \vspace{2mm}
    \\
    Project Page: \href{https://hit-perfect.github.io/VideoMemory/}{VideoMemory}

}
\author{\authorBlock}

\begin{document}
\twocolumn[{
    \renewcommand\twocolumn[1][]{#1}%
    \maketitle
    \begin{center}
        \centering
        \vspace{-2em}
        \includegraphics[width=\linewidth]{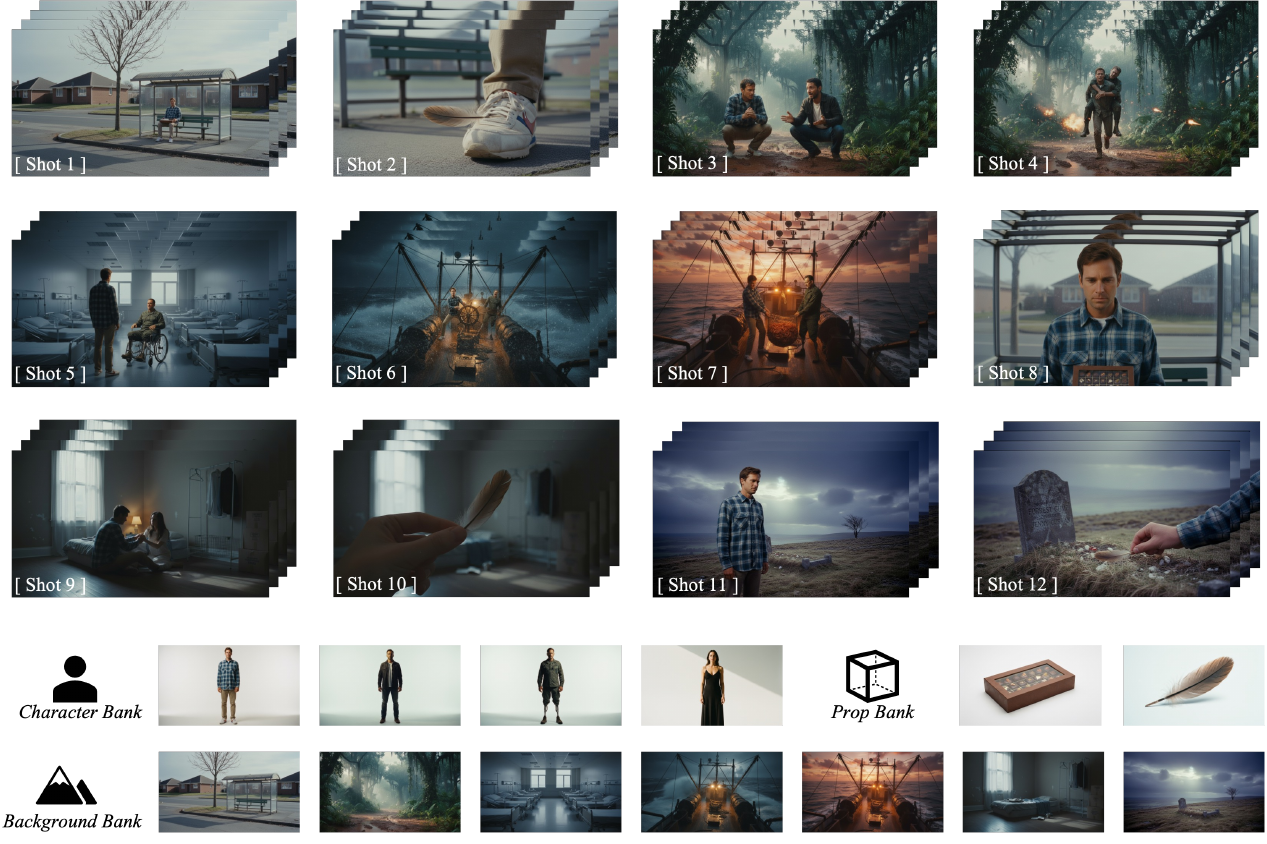}
        \captionof{figure}{From a single story prompt, \texttt{VideoMemory} generates coherent multi-shot videos using dynamic Character, Prop, and Background Memory Banks. This ensures exceptional entity consistency, as seen with the \textbf{feather} prop remaining perfectly stable across distant shots (e.g., 2, 10, and 12) despite significant scene and viewpoint variations.}
        \label{teaser}
    \end{center}
}]
\begingroup
\makeatletter
\renewcommand\thefootnote{}%
\renewcommand\@makefntext[1]{\noindent#1}%
\footnotetext{%
  $^*$ Equal contribution.\\
  $^\dagger$  Project leader.
  
  $^\ddagger$ Corresponding author.
}
\makeatother
\endgroup

\begin{abstract}
Maintaining consistent characters, props, and environments across multiple shots is a central challenge in narrative video generation. Existing models can produce high-quality short clips but often fail to preserve entity identity and appearance when scenes change or when entities reappear after long temporal gaps. We present VideoMemory, an entity-centric framework that integrates narrative planning with visual generation through a Dynamic Memory Bank. Given a structured script, a multi-agent system decomposes the narrative into shots, retrieves entity representations from memory, and synthesizes keyframes and videos conditioned on these retrieved states. The Dynamic Memory Bank stores explicit visual and semantic descriptors for characters, props, and backgrounds, and is updated after each shot to reflect story-driven changes while preserving identity. This retrieval–update mechanism enables consistent portrayal of entities across distant shots and supports coherent long-form generation. To evaluate this setting, we construct a 54-case multi-shot consistency benchmark covering character-, prop-, and background-persistent scenarios. Extensive experiments show that VideoMemory achieves strong entity-level coherence and high perceptual quality across diverse narrative sequences. 
\end{abstract}


\section{Introduction}


Modern text-to-video generation has made rapid progress in recent years, with large-scale diffusion transformers~\cite{ho2020denoising,song2020denoising,rombach2022high} and video generative models~\cite{kong2024hunyuanvideo, wan2025, meituanlongcatteam2025longcatvideotechnicalreport, openai_sora2, deepmind_veo} capable of producing short clips with high visual fidelity, smooth motion, and diverse appearance styles from open-domain prompts. Beyond standalone clips, this progress has sparked growing interest in story-driven video generation, where a concise script is translated into a coherent sequence of scenes and shots that follow a narrative structure and can be directly used for pre-visualization, storyboarding, and content creation. However, realizing such story-level generation imposes far stricter requirements than conventional short video synthesis: main and supporting characters, signature props, and key environments must remain consistent across distant shots under varying viewpoints, lighting conditions, and narrative contexts, while only exhibiting intentional, story-driven changes.


Recent works have explored several directions toward more coherent story-level video generation. 
Plan-then-generate pipelines~\cite{lin2023videodirectorgpt,hu2024storyagent,zheng2024videogenofthought} use large language models to expand a brief prompt into structured scene–shot scripts, providing explicit narrative planning and improving control over the presence and arrangement of entities. 
One-pass multi-shot generation methods~\cite{guo2025long, wang2025echoshot, kara2025shotadapter, meng2025holocine} model multiple shots jointly or sequentially, leveraging extended temporal context to improve cross-shot coherence. 
In parallel, memory-inspired approaches~\cite{zhang2025framepack, yu2025context, wu2025video} manage long-range dependencies by compressing historical frames, aligning viewpoints across time, or maintaining persistent scene-level representations for retrieval. 

Despite this progress, long-range entity consistency remains a core challenge in narrative video generation. Plan-then-generate pipelines provide narrative structure but lack persistent visual representations, causing each shot to be generated without reference to past appearances. One-pass multi-shot generation methods capture broader temporal context, yet their histories are feature-centric and do not preserve explicit entity states, leading to drift when characters or props reappear after long gaps. Memory-inspired approaches store information at the frame or feature level and offer no dedicated slots for named entities, making identity recovery unreliable once an entity leaves the local context window.

\begin{figure}[t]
    \centering
    \includegraphics[width=1.0\linewidth]{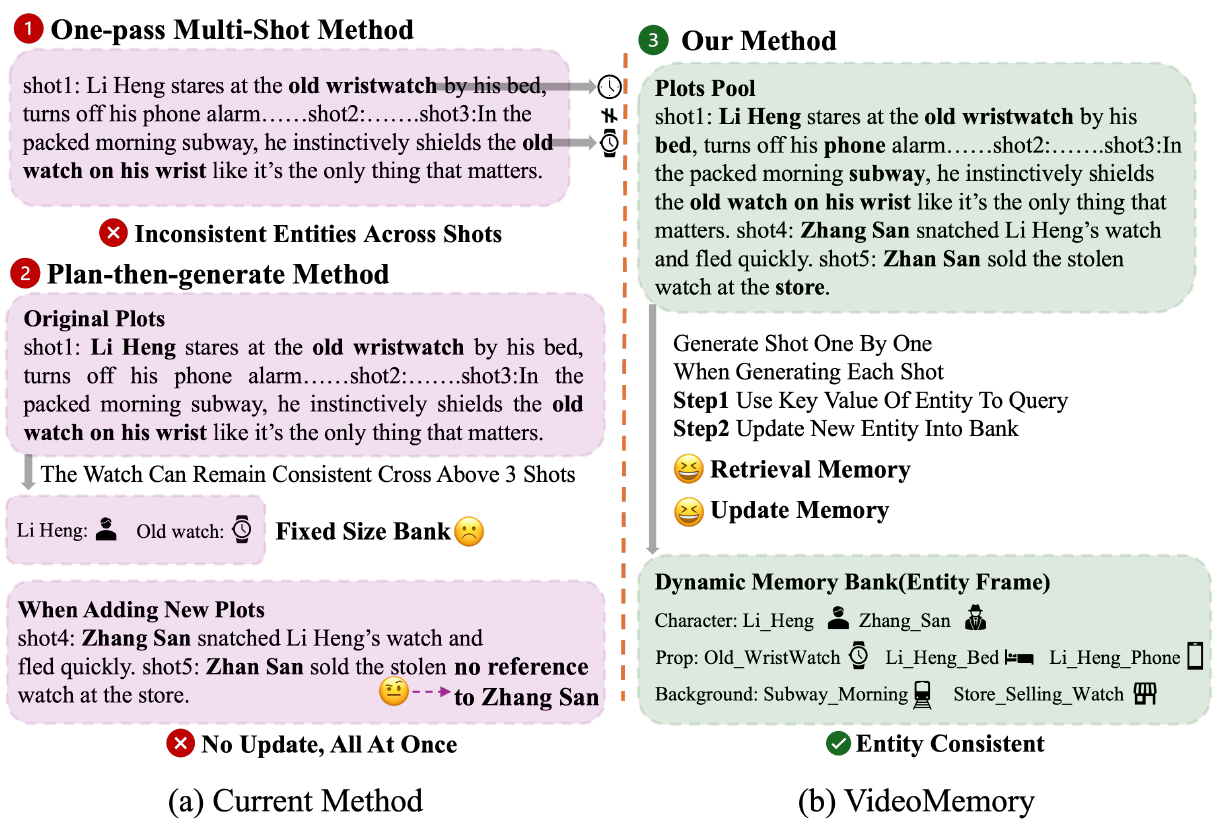}
    \caption{\textbf{Comparison between existing methods and our method.}
(a) Existing methods lack entity-level memory, resulting in inconsistent appearances across shots.
(b) \emph{VideoMemory} retrieves and updates an entity-aware Dynamic Memory Bank during generation, enabling consistent multi-shot videos from text alone.}
    \label{fig:Comparison}
    \vspace{-1.3em}
\end{figure}

We draw inspiration from human production workflows. In real film and television projects~\cite{cleve2006film,honthaner2013complete}, production crews first define a set of story assets, including cast appearances, costumes, makeup, key props, and sets, and then update this inventory as the narrative unfolds so that recurring entities remain recognizable while reflecting plot-driven changes. This motivates an explicit, dynamically updated memory for key entities in story-centric video generation. This motivates an explicit, dynamically updated memory for key entities in story-centric video generation.

In this work, we introduce \emph{VideoMemory}, an entity-centric framework for script-to-multi-shot video generation built around a Dynamic Memory Bank. The Dynamic Memory Bank is an explicit, entity-aware external memory that stores and updates the evolving visual and semantic states of characters, props, and backgrounds, enabling long-range consistency and controlled narrative evolution. Given a short script, a Storyboard Agent expands it into structured scene and shot descriptions. A Memory Agent extracts the entities required for each shot, forms attribute-based keys, and interacts with the Dynamic Memory Bank, which stores explicit visual and semantic representations of characters, props, and backgrounds. Before generating a shot, the agent retrieves the corresponding entity states from the memory. A Visualization Agent then synthesizes keyframes and shot videos conditioned on these retrieved representations to ensure consistent appearance of recurring entities. After each shot is produced, the Memory Agent updates the Dynamic Memory Bank to incorporate narrative changes such as aging, clothing variations, or prop condition while preserving identity. Through this coordinated retrieve-and-update mechanism over a shared external memory, VideoMemory maintains coherent entity evolution across distant shots without relying on full-frame history caching. A qualitative illustration is shown in Figure~\ref{fig:Comparison}. 

To systematically assess entity-level consistency under diverse narrative settings, we further construct a dedicated multi-shot consistency benchmark comprising 54 structured cases that span character-, prop-, and background-persistent scenarios with varying shot lengths. The benchmark provides story-driven test situations that require models to preserve identity while accommodating intentional narrative changes. Extensive experiments on this benchmark demonstrate that VideoMemory achieves strong cross-shot coherence, stable long-range entity continuity, and high perceptual realism across all categories.

Our contributions are three-fold:
\begin{itemize}
    \item We introduce \emph{VideoMemory}, an entity-centric framework for script-to-multi-shot video generation built around a Dynamic Memory Bank that explicitly tracks and updates the visual and semantic states of narrative entities.
    \item We develop a retrieval–and-update memory mechanism with dedicated slots for characters, props, and environments, enabling long-range entity consistency while enabling long-range entity consistency while allowing story-guided changes in their visual state.
    \item We construct a  multi-shot consistency benchmark with 54 story-driven cases covering character-, prop-, and background-persistent scenarios, providing the first structured evaluation protocol for long-range entity coherence in narrative videos.
\end{itemize}  
\section{Related Work}

\subsection{Story-Driven Video Generation}
Storyboard visualization aims to automatically generate visual imagery or videos from textual or conceptual inputs, serving as a crucial component in story planning and cinematic pre-visualization. With the rapid progress of generative models, this area has seen significant improvements in both visual quality and narrative coherence.

A common line of work adopts a keyframe-first strategy, where representative frames are first synthesized and then expanded into complete videos using image-to-video (I2V) generation models. These approaches \cite{zhou2024storydiffusion, li2018storygan, ma2024magic, wei2023dreamvideo, gu2025roictrl, huang2024context} focus on ensuring character consistency across shots, improving continuity between generated keyframes. Building upon this, several recent methods\cite{lin2023videodirectorgpt, long2024videostudio, xie2024dreamfactory, hu2024storyagent, zheng2024videogenofthought, wu2025automated, huang2025filmaster} integrate large language models (LLMs) to better control and plan narrative elements, thereby enhancing consistency in story visualization.

Another emerging direction is multi-shot generation\cite{guo2025long, wang2025echoshot, kara2025shotadapter, wang2025storyanchors}, in which models are trained to generate multiple consecutive shots within a single forward pass. This is typically achieved by designing specialized architectures and collecting large-scale datasets of cinematic transitions. 

Although these approaches improve temporal smoothness, they still struggle to achieve object-level consistency, particularly in re-creating previously appeared entities or props with stable attributes.To address these limitations, we propose a dynamic Memory Bank mechanism that stores previously generated entities according to their semantic attributes and key descriptors. This stored memory can be efficiently retrieved and updated during subsequent generation, enabling more coherent and faithful reproduction of characters and objects across shots.

\subsection{Memory in Video Generation}
Recent advances in video generation have achieved remarkable progress in both visual quality and temporal coherence. Most contemporary approaches are built upon the Diffusion Transformer framework\cite{peebles2023scalable}, which has become the dominant paradigm in this field. Open-source models such as HunyuanVideo\cite{kong2024hunyuanvideo}, Wan 2.1\&2.2\cite{wan2025} demonstrate impressive performance for short video generation, producing high-fidelity and temporally stable results within limited frame ranges. However, these models encounter severe challenges when extrapolating to long-duration videos, often exhibiting repetitive patterns and content drift\cite{zhao2025riflex}.

To overcome this limitation, several studies\cite{teng2025magi, chen2025skyreels, huang2025self, yang2025longlive, dalal2025one, cui2025self, liu2025rolling} have explored long-video generation. While these methods successfully extend video duration, they still fail to preserve object reappearance consistency—the ability to maintain identical appearance and structure of an object that reoccurs after a long temporal gap. This phenomenon reveals an intrinsic memory degradation problem in diffusion-based video models.

To mitigate memory loss, FramePack\cite{zhang2025framepack} compresses contextual representations so that the model can reference previous content during generation. Context-as-Memory\cite{yu2025context} introduces field-of-view (FOV) alignment to match camera perspectives across frames, and \cite{wu2025video} maintains a dynamically updated 3D point-cloud memory to provide scene-level references. Although these methods advance the understanding of memory mechanisms in video generation, they still suffer from high maintenance cost and limited consistency, particularly in preserving the geometric integrity of objects across frames.

To address these issues, we propose a Memory Bank framework that introduces explicit memory retrieval and update modules for managing key characters and objects. By storing only critical entities and updating them efficiently, our approach reduces memory overhead and substantially improves long-term consistency in both appearance and structure.

\section{Method}
\label{sec:method}


\subsection{Task Formulation}
\label{subsec:task_formulation}

Given a script synopsis $S$, the goal of automated video generation is to produce
a multi-shot video $\hat{V}$ that is not only visually plausible but also
narratively coherent and temporally consistent. In particular, the generated
video should preserve character identity across time, maintain the appearance
and state of key props, and keep background settings compatible with the story
progression. Formally, we seek an automated mapping
\begin{equation}
    \mathcal{F}: S \rightarrow \hat{V},
\end{equation}
where $\hat{V} = \{\hat{V}_i \mid i = 1, 2, \ldots, N\}$ and $\hat{V}_i$
denotes the $i$-th shot.  

The function $\mathcal{F}(\cdot)$ implicitly encompasses two tightly coupled
parts. The first part performs high-level planning: it decomposes the synopsis
into sub-scripts, scenes, and shots, and decides camera-related parameters such
as scale, perspective, and basic motion patterns. The second part is
responsible for visual realization: it materializes the planned content into
keyframes and shot-level videos while respecting long-range constraints on
characters, props, and environments. In \emph{VideoMemory}, $\mathcal{F}$ is
implemented as a memory-guided, multi-agent system that operates in a
shot-by-shot manner and explicitly maintains a Dynamic Memory Bank for
entity-level consistency.

\subsection{Multi-Agent Cooperation}
\label{subsec:multi_agent}

\begin{figure*}[t]
    \centering
    \includegraphics[width=1.0\linewidth]{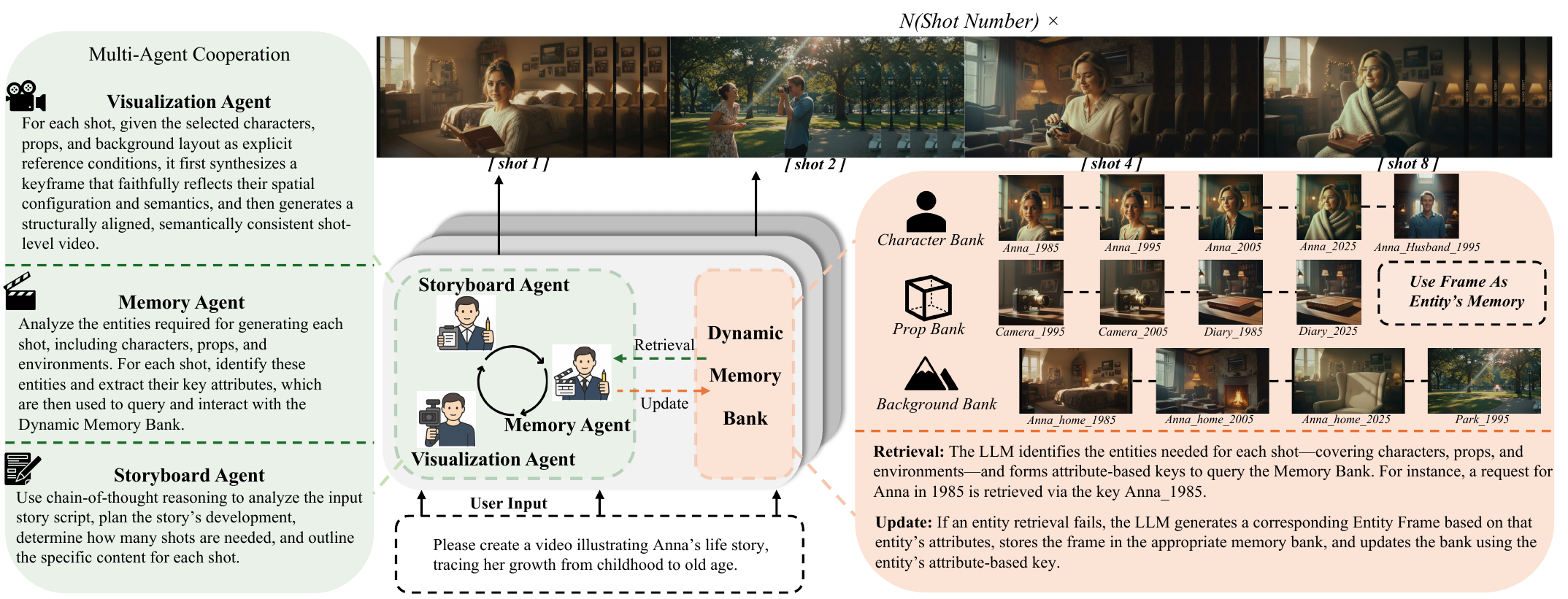}
    \caption{\textbf{The framework of the proposed \emph{VideoMemory}.}
    Starting from a script synopsis, our system plans shot-level descriptions,
    interacts with a Dynamic Memory Bank to retrieve or create entity references,
    generates keyframes, and finally synthesizes a coherent multi-shot video.}
    \label{fig:Framework}
    \vspace{-1.3em}
\end{figure*}

To realize the mapping $\mathcal{F}$, we decompose the generation process into
three cooperating agents (Fig.~\ref{fig:Framework}): a \emph{Storyboard Agent},
a \emph{Memory Agent}, and a \emph{Visualization Agent}. Each agent is
responsible for a specific stage of the pipeline and exposes clean interfaces
to the others. The Storyboard Agent converts the high-level synopsis into
structured shot descriptions. The Memory Agent analyzes each shot, identifies
the required entities, and interacts with the Dynamic Memory Bank to retrieve
or create reference images. The Visualization Agent uses these references to
generate keyframes and then delegates motion synthesis to an image-to-video
(I2V) model. The whole procedure is summarized in
Algorithm~\ref{alg:videomemory}.


\begin{algorithm}[t]
\small
\caption{\textbf{VideoMemory}: Procedure from Script Synopsis to Multi-Shot Video}
\label{alg:videomemory}
\begin{algorithmic}[1]
\Require Script synopsis $S$
\Ensure Multi-shot video $\hat{V} = \{\hat{V}_i\}_{i=1}^{N}$
\State $\{\mathcal{C}_i\}_{i=1}^{N} \gets \textsc{StoryboardAgent}(S)$
\State Initialize memory banks $\mathcal{M}^{\text{char}}, \mathcal{M}^{\text{prop}}, \mathcal{M}^{\text{bg}} \gets \emptyset$
\For{shot $i = 1$ to $N$} \Comment{Process shots sequentially}
    \State $\{(e_{ij}, {a}_{ij}, c_{ij})\}_j \gets \textsc{MemoryAgentAnalyze}(\mathcal{C}_i)$
    \For{each $(e_{ij}, {a}_{ij}, c_{ij})$}
        \State $(I_{ij}^{\text{ref}}, \mathcal{M}) \gets \textsc{RetrieveOrGenerate}(e_{ij}, {a}_{ij}, c_{ij}, \mathcal{M})$
    \EndFor
    \State $I_i^{\text{key}} \gets \textsc{VisualizationAgentKeyframe}(\mathcal{C}_i,\{I_{ij}^{\text{ref}}\}_j)$
    \State $\hat{V}_i \gets \textsc{I2VModel}(I_i^{\text{key}}, \mathcal{C}_i)$
\EndFor
\State $\hat{V} \gets \{\hat{V}_i\}_{i=1}^{N}$
\end{algorithmic}
\end{algorithm}

\paragraph{Storyboard Agent.}
The Storyboard Agent acts as the \emph{director} of the pipeline. Given the
synopsis $S$, it uses a large language model (LLM) with chain-of-thought
reasoning to unfold the high-level story into a detailed narrative and a
sequence of structured shots. Concretely, the agent first segments the story
into major beats (e.g., “prologue in the village”, “conflict on the bridge”),
then refines each beat into one or several shots by specifying the spatial
setup, the participating characters, and the key events that should happen on
screen. The final output is a set of shot-level descriptions
$\{\mathcal{C}_i\}_{i=1}^{N}$, where each $\mathcal{C}_i$ contains textual
information such as camera distance, framing (e.g., close-up vs.\ wide shot),
and the temporal order of actions. This stage defines \emph{what} should be
shown in each shot and provides the narrative scaffold for all downstream
modules.

\paragraph{Memory Agent.}
The Memory Agent plays the role of a \emph{casting and prop specialist}. Given
a shot description $\mathcal{C}_i$, it first parses the text to identify all
entities that must appear in the shot, including characters, key props, and
background elements. For each detected entity $e_{ij}$, $j$ represents the j-th 
entity of this shot, the agent constructs an attribute vector $a_{ij}$ that summarizes its semantic state in the
current shot, such as age, hairstyle, outfit, emotional state, and time
period. Entities are also tagged by a category label
$c_{ij} \in \{\text{char}, \text{prop}, \text{bg}\}$ so that they can be routed
to the correct memory bank.

Once the entity set $\{(e_{ij}, a_{ij}, c_{ij})\}_j$ is obtained, the
Memory Agent queries the Dynamic Memory Bank (Sec.~\ref{subsec:memory_bank})
for each entity and either retrieves an existing reference image or triggers
the creation of a new one. The retrieved or newly generated images
$\{I_{ij}^{\text{ref}}\}_j$ are then passed to the Visualization Agent. By
centralizing all entity analysis and memory interaction within this agent, we
ensure that consistency decisions are made explicitly and can be inspected or
visualized.

\paragraph{Visualization Agent.}
The Visualization Agent plays the role of a \emph{cinematographer} and is
responsible for turning symbolic specifications into pixels. For each shot $i$,
it receives the shot description $\mathcal{C}_i$ and the retrieved entity
reference images $\{I_{ij}^{\text{ref}}\}_j$ from the Memory Agent. Based on
these inputs, it first composes a shot-level keyframe that encodes the desired
composition, character poses, and background layout while remaining consistent
with the memorized appearance of all entities. It then uses this keyframe,
together with the same shot description $\mathcal{C}_i$, to drive an
image-to-video (I2V) model and produce the corresponding shot video. The
keyframe construction and video synthesis steps are detailed and formalized in
Sec.~\ref{subsec:keyframe_video}.

\begin{figure*}[t]
    \centering
    \includegraphics[width=1.0\linewidth]{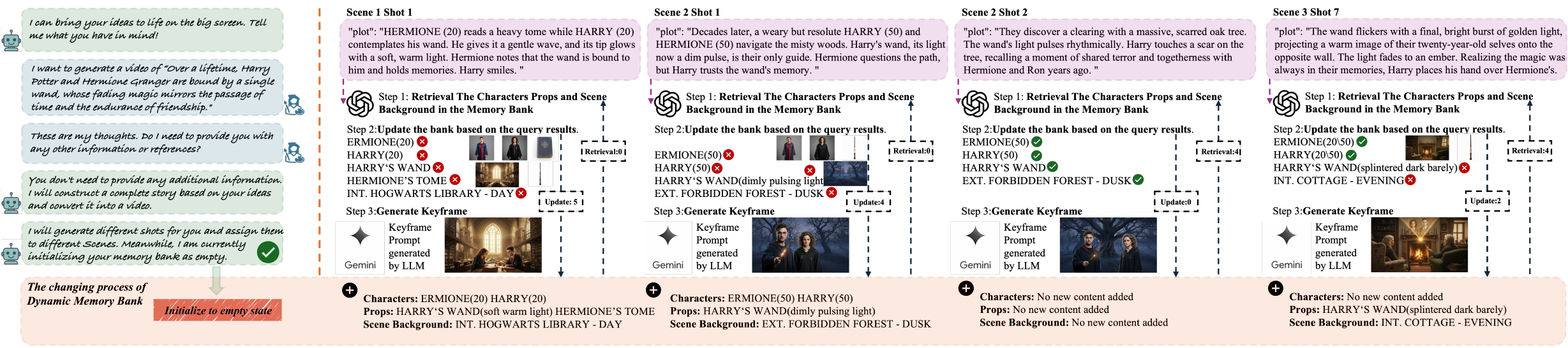}
    \caption{\textbf{An example ``Harry Potter'' workflow of \emph{VideoMemory}.}
    Starting from a textual synopsis, VideoMemory plans all scenes and shots
    without any visual input, incrementally builds the Dynamic Memory Bank while
    generating each shot, and finally composes a multi-shot sequence in which
    characters, props, and environments remain consistent over time (e.g., in Shot~7,
    Harry and Hermione at different ages are rendered coherently together with that shot's
    props and background).}
    \label{fig:Case_Harry_Potter}
    \vspace{-1.3em}
\end{figure*}

\subsection{Dynamic Memory Bank}
\label{subsec:memory_bank}
The Dynamic Memory Bank is the central component that enforces entity-level
consistency throughout the video. Instead of sampling a new appearance for each
character, prop, or background region in every shot, \emph{VideoMemory} stores
entity-specific reference images and reuses or adapts them when the same
entity reappears. This mechanism not only stabilizes character identity and
object appearance, but also keeps recurring environments (e.g., a classroom, a
street corner, or a castle hall) structurally and stylistically aligned across
shots. Such design is crucial for long videos where characters, props, and
scenes can undergo significant temporal changes (e.g., aging, damage,
re-decoration, day–night transitions) while still remaining visually
recognizable and coherent at the story level.

\paragraph{Memory representation.}
We maintain three specialized banks for different semantic roles: character
memory $\mathcal{M}^{\text{char}}$, prop memory $\mathcal{M}^{\text{prop}}$,
and background memory $\mathcal{M}^{\text{bg}}$. Each entry in these banks is
an \emph{entity frame}
\[
    m = (e, {a}, I^{\text{ref}}),
\]
where $e$ is the entity identifier (e.g., \emph{Anna}), ${a}$ is an
attribute vector encoding its semantic state, and $I^{\text{ref}}$ is a
reference image that visually realizes this state. In practice, we store
$e$ and ${a}$ as text and index them with an LLM-friendly format (e.g.,
“Anna\_1985”), while $I^{\text{ref}}$ is stored as a PNG file on
disk. This representation makes it straightforward to inspect and reuse past
appearances in later shots.

\paragraph{Retrieval.}
Given an entity $(e_{ij}, {a}_{ij}, c_{ij})$ from shot $i$, the Memory
Agent first selects the corresponding bank according to $c_{ij}$ and then
searches for the most semantically compatible entry. Rather than relying on
exact string matching, we compute semantic similarity between the current
attributes and stored attributes using an LLM-based matcher. This allows the
system to retrieve appropriate references even when the textual descriptions
are phrased differently but share the same meaning, such as “middle-aged Anna
in a winter coat” and “Anna, about 45, wearing a thick coat in the snow”.  
If a suitable entry is found, its reference image $I^{\text{ref}}$ is returned
as $I_{ij}^{\text{ref}}$ and reused directly in the current shot.

\paragraph{Update.}
If retrieval fails (i.e., no sufficiently similar entry exists), the Memory
Agent calls the image generation model to create a new reference:
\[
    I_{ij}^{\text{new}} = \mathcal{G}(e_{ij}, {a}_{ij}, \mathcal{H}_{e_{ij}}),
\]
where $\mathcal{G}(\cdot)$ is implemented by a image generator, and $\mathcal{H}_{e_{ij}}$ denotes the set of
historical reference images for entity $e_{ij}$, if any. When
$\mathcal{H}_{e_{ij}}$ is non-empty (for example, when generating “Anna at age
60” after having already generated “Anna at age 20”), the model is prompted
with both the current attributes and the past images so that it can produce a
new appearance that reflects the temporal change while preserving core
identity features. The new triplet $(e_{ij}, {a}_{ij}, I_{ij}^{\text{new}})$
is then inserted into the corresponding bank. By repeatedly applying this
retrieve-or-update procedure across all shots, the Dynamic Memory Bank builds
a rich, temporally ordered collection of entity states that supports both
consistency and controlled evolution.

\subsection{Keyframe \& Video Generation}
\label{subsec:keyframe_video}

Based on the shot descriptions $\{\mathcal{C}_i\}$ and the entity reference
images $\{I_{ij}^{\text{ref}}\}$ prepared by the previous agents, it converts
symbolic specifications into pixel-level videos in two steps.

For each shot $i$, the agent first synthesizes a shot-level keyframe that
reflects both the desired composition and the memorized appearance of all
entities:
\begin{equation}
    I_i^{\text{key}} =
    \mathcal{A}_{\text{vis}}\big(\mathcal{C}_i,\{I_{ij}^{\text{ref}}\}_j\big),
\end{equation}
where $\mathcal{A}_{\text{vis}}$ denotes an image generation module explicitly
conditioned on the entity references. These keyframes act as visually
consistent anchors for subsequent video synthesis.

Next, the keyframe $I_i^{\text{key}}$ and its shot description $\mathcal{C}_i$
are fed into an image-to-video (I2V) video generation model, such as Wan~2.2, to
produce the corresponding shot video:
\begin{equation}
    \hat{V}_i =
    \mathcal{A}_{\text{i2v}}\big(I_i^{\text{key}},\, \mathcal{C}_i\big),
\end{equation}
where $\mathcal{A}_{\text{i2v}}$ denotes the I2V generation function. The model
adds motion and camera dynamics while remaining faithful to the spatial
structure and entity appearances specified in the keyframe.

Finally, the complete video is obtained by concatenating all generated shots in
narrative order:
\begin{equation}
    \hat{V} = \{\hat{V}_i\}_{i=1}^{N}.
\end{equation}
Because every keyframe is generated under the guidance of the same Dynamic
Memory Bank and then expanded by the I2V model, the resulting multi-shot video
exhibits stable character identities, persistent props, and coherent
environments over long temporal ranges.

A concrete end-to-end example on a ''Harry and Hermione'' story is illustrated in Fig.~\ref{fig:Case_Harry_Potter}.

\section{Experiments}
\label{sec:experiments}

\begin{table*}[t]
\centering

\caption{\textbf{Multi-shot consistency results.}
We evaluate character, prop, and background consistency using DINOv2 similarity.
Our method achieves superior performance across all metrics, especially as the
number of shots increases. Best and second-best scores are marked in
\textbf{bold} and \underline{underlined}, respectively.}
\label{tab:quantitative_results}
\vspace{-0.4em}
\setlength{\tabcolsep}{8pt}
\resizebox{\textwidth}{!}{%
\begin{tabular}{l *{12}{C}}
\toprule
 & \multicolumn{4}{c}{Character Consistency$\uparrow$} & \multicolumn{4}{c}{Prop Consistency$\uparrow$} & \multicolumn{4}{c}{Background Consistency$\uparrow$} \\
\cmidrule(lr){2-5} \cmidrule(lr){6-9} \cmidrule(lr){10-13}
Shot Number & 4 & 8 & 12 & Avg. & 4 & 8 & 12 & Avg. & 4 & 8 & 12 & Avg. \\
\midrule
Wan2.2~\cite{wan2025} & 0.34 & - & - & - & 0.48 & - & - & - & 0.25 & - & - & - \\
EchoShot~\cite{wang2025echoshot} & 0.45 & 0.44 & - & - & 0.59 & \textbf{0.51} & - & - & 0.54 & 0.37 & - & - \\
IC-LoRA+Wan2.2~\cite{huang2024context} & 0.42 & 0.55 & 0.43 & 0.47 & 0.50 & 0.44 & 0.34 & 0.43 & 0.31 & 0.33 & 0.29 & 0.31 \\
StoryDiffusion+Wan2.2~\cite{zhou2024storydiffusion} & 0.53 & 0.62 & 0.46 & 0.54 & 0.43 & 0.47 & \underline{0.52} & 0.47 & 0.51 & 0.40 & 0.36 & 0.42 \\
VGoT+Wan2.2~\cite{zheng2024videogenofthought} & \underline{0.59} & 0.53 & \underline{0.60} & \underline{0.57} & \underline{0.48} & 0.22 & 0.24 & 0.31 & \underline{0.53} & \underline{0.36} & \underline{0.47} & \underline{0.45} \\
\midrule
VideoMemory (Ours) & \textbf{0.61} & \textbf{0.65} & \textbf{0.64} & \textbf{0.63} & \textbf{0.69} & \underline{0.50} & \textbf{0.55} & \textbf{0.58} & \textbf{0.71} & \textbf{0.72} & \textbf{0.73} & \textbf{0.72} \\
\bottomrule
\end{tabular}%
}
\end{table*}

\begin{table*}[t]
\centering
\caption{\textbf{User study results.}
We conduct pairwise comparisons between \emph{VideoMemory} and each baseline
under four consistency dimensions. \textbf{Numbers denote the percentage of trials in
which human experts preferred the video generated by \emph{VideoMemory}}.}
\label{tab:user_study}
\vspace{-0.4em}
\setlength{\tabcolsep}{8pt}
\resizebox{\textwidth}{!}{%
\begin{tabular}{lcccc}
\toprule
\textbf{Baseline} & Character Consistency & Prop Consistency & Background Consistency & Semantic Consistency \\
\midrule
Wan2.2~\cite{wan2025} & 87.50\% & 79.17\% & 87.50\% & 94.44\% \\
EchoShot~\cite{wang2025echoshot} & 91.66\% & 91.66\% & 70.83\% & 91.66\% \\
IC-LoRA+Wan2.2~\cite{huang2024context} & 95.83\% & 87.50\% & 91.66\% & 95.83\% \\
StoryDiffusion+Wan2.2~\cite{zhou2024storydiffusion} & 91.66\% & 79.17\% & 91.66\% & 87.50\% \\
VGoT+Wan2.2~\cite{zheng2024videogenofthought} & 95.83\% & 83.33\% & 95.83\% & 91.66\% \\
\bottomrule
\end{tabular}%
}
\end{table*}

\subsection{Experimental Setup}

\paragraph{Implementation Details.}
We implement our multi-agent framework using Gemini 2.5 Pro~\cite{deepmind_gemini_pro} as the core LLM for all narrative reasoning tasks, including script parsing, entity attribute extraction, and structured prompt generation. For visual synthesis, keyframes are generated with Gemini 2.5 Flash-Image~\cite{google_gemini_flash_image} and then animated into video clips using the open-source Wan 2.2 I2V-14B model~\cite{wan2025}. Unless otherwise specified, all experiments use this fixed backbone, so that performance differences can be attributed primarily to our memory mechanism rather than to changes in generative capacity.

\paragraph{Evaluation Metrics.}
To assess cross-shot consistency, we use three complementary metrics computed on the middle frame of each shot and averaged over all benchmark cases:
(a) Character Consistency: DINOv2~\cite{oquab2023dinov2} feature similarity over all detected face regions, localized with Mediapipe~\cite{lugaresi2019mediapipeframeworkbuildingperception}, quantifying the stability of character identity.
(b) Prop Consistency: DINOv2 similarity on target prop regions, identified using Grounded SAM~\cite{ren2024grounded} conditioned on textual prop descriptions, measuring appearance stability of the object.
(c) Background Consistency: DINOv2 similarity between full frames (without segmentation), evaluating stability of the scene environment under camera and narrative changes.

\paragraph{Benchmark for Multi-Shot Entity Consistency.}
Existing video benchmarks largely target single-shot or short clips and rarely provide structured narratives, making them ill-suited for evaluating long-horizon entity persistence. We therefore construct a benchmark of 54 test cases, structured as $3$ subclasses $\times$ $3$ shot lengths $\times$ $6$ samples. The subclasses isolate a single consistent factor: Character-persistent, Prop-persistent, or Background-persistent, evaluated at $K \in \{4, 8, 12\}$ shots. In each subclass, the two non-persistent factors are deliberately varied across shots; e.g., in Character-persistent cases the same character must persist while scene and props change, forcing models to decouple the target entity from its context. Each case forms a coherent $K$-shot narrative with a three-act structure. Prompt construction and full evaluation protocols are detailed in the Supplementary Material.

\paragraph{Baselines.}
We compare \textit{VideoMemory} to a diverse set of representative baselines: (1) native text-to-video (T2V) models Wan 2.2 T2V-14B~\cite{wan2025} and EchoShot~\cite{wang2025echoshot}, evaluated in a one-pass setting to probe their intrinsic consistency; and (2) two-stage, keyframe-first methods IC-LoRA~\cite{huang2024context}, StoryDiffusion~\cite{zhou2024storydiffusion}, and VGoT~\cite{zheng2024videogenofthought}. All keyframe-first methods are paired with the same Wan 2.2 I2V-14B model for the final animation stage, and all methods receive the same set of $K$ per-shot textual descriptions.

\begin{figure*}[t]
    \centering
    \includegraphics[width=1.0\linewidth]{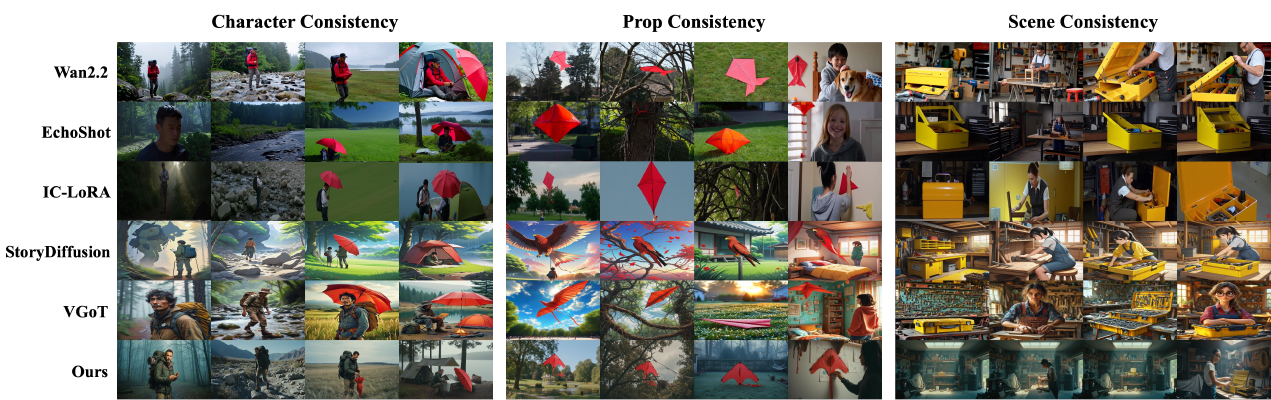}
    \caption{\textbf{Qualitative comparison demonstrating superior entity consistency.} Across all three subclasses (Character, Prop, Background), \textit{VideoMemory} (bottom row) maintains remarkable stability where baselines fail. Note how baselines exhibit severe \textbf{identity drift}—changing a character's appearance (left), morphing a red kite into other objects (middle), and altering a garage's layout (right). In contrast, our method preserves the identity of all entities across distant shots, a direct result of our explicit memory management.}
    \label{fig:qualitative_analysis}
    \vspace{-0.2em}
\end{figure*}

\subsection{Quantitative Results}
\paragraph{Main Results.} As shown in Table~\ref{tab:quantitative_results}, \textit{VideoMemory} establishes a new state-of-the-art on multi-shot entity consistency. It outperforms prior methods across all three metrics---character, prop, and background---with a single exception in a specific setting, and its performance advantage becomes more pronounced as sequence length increases. The aggregated average scores further confirm its superior overall stability. Some baselines could not produce videos at the required shot counts for longer sequence settings, resulting in incomplete scores. We attribute our method's robust long-horizon coherence to the architectural design, where the Dynamic Memory Bank plays a crucial role, a finding substantiated by our ablation study (Table~\ref{tab:ablation_results}).

\paragraph{User Study.}  To complement the quantitative metrics, we conduct a user study with 5 participants. Each participant was presented with 270 randomly video pairs, covering all 54 benchmark cases compared against each of the 5 baselines. For each pair, experts performed a forced-choice task, answering one of three possible \textit{subclass-specific} consistency questions (e.g., “Which better preserves character identity across all shots?”). As reported in Table~\ref{tab:user_study}, our approach was strongly preferred over all baselines, achieving preference rates as high as 95.8
\% for character and background consistency. These results confirm that the significant gains observed in our quantitative metrics translate to a perceptibly more coherent viewing experience.

\subsection{Qualitative Analysis}

Figure~\ref{fig:qualitative_analysis} visually corroborates these trends. Across all three subclasses, baselines often exhibit identity drift, while \textit{VideoMemory} maintains much more stable entities.
\textbf{Character Consistency (left):} Baselines fail to preserve the same character as the scene evolves: facial structure, apparent age, and even ethnicity may change between shots, whereas our method keeps a consistent identity.
\textbf{Prop Consistency (middle):} The “bright RED KITE” is frequently altered by baselines, morphing into other red objects or disappearing. \textit{VideoMemory} preserves the kite’s distinct shape and color across shots.
\textbf{Background Consistency (right):} In Background-persistent cases, the “cluttered garage” background in baseline outputs changes layout; tools and shelves rearrange across shots. Our method produces a more coherent and stable spatial configuration.
These examples illustrate how explicit entity-centric memory helps prevent the types of drift that are visually most disruptive to multi-shot narratives.

\begin{table}[t]
\centering
\caption{\textbf{Ablation study on the Dynamic Memory Bank (DMB). }Removing memory leads to a clear drop in all consistency metrics. Results are averaged over all 8-shot cases.}
\label{tab:ablation_results}
\setlength{\tabcolsep}{6pt}
\begin{tabular}{lcc}
\toprule
Method & w/o DMB & \textbf{Ours} \\
\midrule
Character Consistency $\uparrow$ & 0.51 & \textbf{0.65} \\
Prop Consistency $\uparrow$ & 0.43 & \textbf{0.50} \\
Background Consistency $\uparrow$ & 0.28 & \textbf{0.72} \\
\bottomrule
\end{tabular}
\end{table}

\vspace{-0.4em}
\subsection{Ablation Studies}
\begin{figure}[t]
    \centering
    \includegraphics[width=1.0\linewidth]{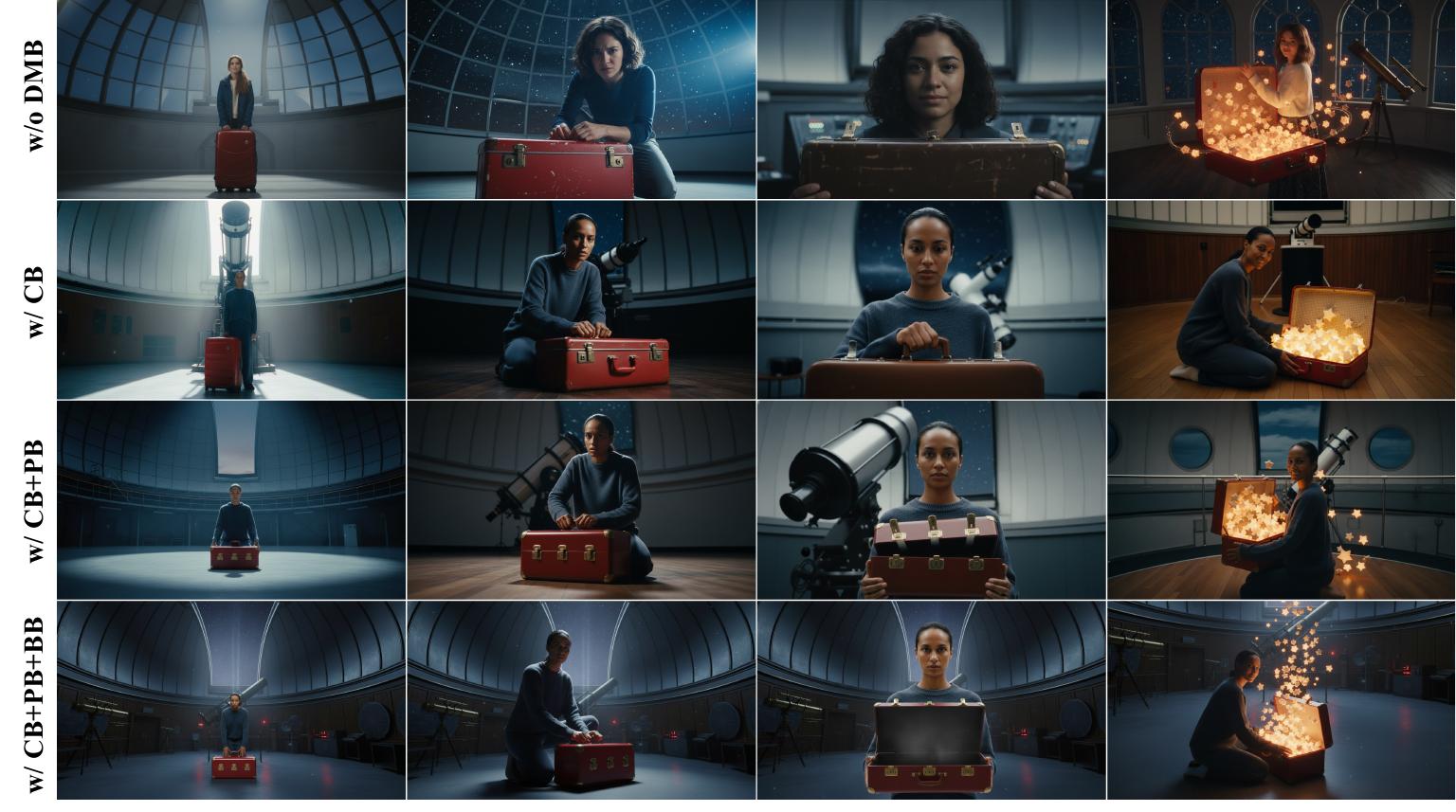}
    
    \caption{\textbf{Qualitative ablation of the memory modules.}
    Top (w/o DMB): The baseline without memory shows noticeable entity drift.
    Rows 2--4: Adding the Character Bank (+CB), Prop Bank (+PB), and Background Bank (+BB) progressively stabilizes the person, the red suitcase, and the observatory. The bottom row shows the full model.}
    \label{fig:ablation}
\vspace{-0.4em}
    
\end{figure}

To validate the efficacy of our Dynamic Memory Bank (DMB), we conducted an ablation study. As shown in Table~\ref{tab:ablation_results}, removing the DMB entirely (\textbf{w/o DMB}) causes performance to collapse across all entity consistency metrics, with background consistency suffering a catastrophic drop from 0.72 to 0.28. Figure~\ref{fig:ablation} qualitatively corroborates this, showing severe drift for the character, prop, and background in the absence of memory. Sequentially re-introducing the Character Bank (\textbf{+CB}), Prop Bank (\textbf{+PB}), and Background Bank (\textbf{+BB}) incrementally resolves each corresponding inconsistency, culminating in the holistically stable output of our full model. This dual quantitative and qualitative evidence confirms that the DMB is the central driver of consistency and that each of its components is essential.

\section{Conclusion}

We presented \emph{VideoMemory}, an entity-centric framework for script-to-multi-shot video generation that employs a Dynamic Memory Bank to explicitly maintain and update the visual and semantic states of narrative entities across shots. Integrated with a multi-agent pipeline for narrative decomposition and visual synthesis, this memory-based formulation supports coherent long-range entity continuity while accommodating story-driven variations. To enable systematic evaluation, we introduced a 54-case benchmark targeting character-, prop-, and background-level consistency. Empirical results demonstrate that \emph{VideoMemory} delivers robust entity coherence and high perceptual quality across diverse narrative settings, highlighting the potential of explicit entity memory  for future story-level video generation.



\maketitlesupplementary

\appendix

\section{Full Videos}
For complete videos, please refer to the \href{https://hit-perfect.github.io/VideoMemory/}{project page}.

\section{Evaluation Metrics}
To assess cross-shot consistency in a factorized and interpretable manner, we design three complementary metrics, one for each benchmark subclass (Character-persistent, Prop-persistent, Background-persistent). All metrics operate on the middle frame of each shot and share the same reference-based protocol; they differ only in how visual features are extracted. For a given subclass and shot length $K$, we first compute a per-sequence consistency score and then average these scores over all benchmark cases in that subset. Unless otherwise noted, Character / Prop / Background Consistency are always reported on the Character- / Prop- / Background-persistent subsets, respectively.

For a sequence whose script specifies $N_{\text{req}}$ shots, we first take the middle frame of the \emph{first} shot as a reference frame and extract the corresponding visual features. For every subsequent shot $i$, we extract features from its middle frame and compute the cosine similarity to the reference features. The per-sequence consistency score is obtained by aggregating these similarities using the normalization rule described below. The final metric is the average over all benchmark cases in the corresponding subclass and shot-length setting.

\subsection{Character Consistency}
This metric measures how well the identity of the main character is preserved across shots in the Character-persistent benchmark cases. For each middle frame, we use Mediapipe~\cite{lugaresi2019mediapipeframeworkbuildingperception} to detect all visible faces and crop the corresponding regions. We then extract DINOv2~\cite{oquab2023dinov2} features from these face crops and aggregate them into a single descriptor per frame. Let $\mathbf{f}^{\text{char}}_{1}$ denote the character descriptor of the reference frame (the first shot), and $\mathbf{f}^{\text{char}}_{i}$ the descriptor of shot $i$. We compute the cosine similarity
$\text{sim}(\mathbf{f}^{\text{char}}_{1}, \mathbf{f}^{\text{char}}_{i})$ for every subsequent shot and aggregate these values according to the normalization rule described below. A higher score indicates that faces across shots are consistently recognized as the same identity, while remaining robust to changes in pose, expression, and local illumination.

\subsection{Prop Consistency}
Prop consistency evaluates the appearance stability of a designated object throughout the sequence for Prop-persistent benchmark cases. Each prop-persistent benchmark case is associated with a textual description of its target prop. Conditioned on this description, we run Grounded SAM~\cite{ren2024grounded} on each middle frame to localize the prop region. From the detected region, we extract DINOv2 features to obtain a prop descriptor $\mathbf{f}^{\text{prop}}_{i}$ for shot $i$ (and $\mathbf{f}^{\text{prop}}_{1}$ for the reference shot). We then compute $\text{sim}(\mathbf{f}^{\text{prop}}_{1}, \mathbf{f}^{\text{prop}}_{i})$ for all shots where the prop is successfully detected, and aggregate them using the same normalization strategy. This metric captures whether the object maintains a coherent visual appearance (e.g., color, shape, texture), even under changes in camera viewpoint or surrounding context.

\subsection{Background Consistency}
To quantify the stability of the scene environment under camera and narrative changes in Background-persistent benchmark cases, we define a background consistency metric. In contrast to the two factor-specific metrics above, we do not perform any spatial segmentation; instead, we extract DINOv2 features from the full middle frame of each shot, yielding descriptors $\mathbf{f}^{\text{bg}}_{i}$. We then compute $\text{sim}(\mathbf{f}^{\text{bg}}_{1}, \mathbf{f}^{\text{bg}}_{i})$ between the reference frame and every subsequent shot and aggregate them as below. This metric reflects how consistently the global background---including layout, large-scale structures, and overall lighting---is preserved in sequences where the environment is designed to remain fixed.

\subsection{Handling Variable Shot Counts}
Some methods (e.g., IC-LoRA) may generate a number of shots $N_{\text{out}}$ that does not exactly match the script requirement $N_{\text{req}}$. We handle such cases in a unified way for all three consistency metrics and all benchmark subclasses. When $N_{\text{out}} > N_{\text{req}}$, we simply discard the extra shots and compute similarities only for the first $N_{\text{req}}$ shots, since the scripts specify exactly $N_{\text{req}}$ shots and we are interested in consistency within this prescribed window. When $N_{\text{out}} < N_{\text{req}}$, we compute similarities between the reference frame and all available subsequent shots (i.e., $N_{\text{out}}-1$ pairs), but normalize by the requested number of non-reference shots $N_{\text{req}}-1$:
\[
\text{score} = \frac{1}{N_{\text{req}} - 1} \sum_{i=2}^{N_{\text{out}}} \text{sim}(\mathbf{f}_{1}, \mathbf{f}_{i}).
\]
This effectively treats missing shots as contributing zero similarity and thus penalizes methods that under-generate. For example, if the script requires $N_{\text{req}} = 8$ shots but a model only produces $N_{\text{out}} = 6$, we accumulate 5 similarity values (from shots 2--6) and divide by 7.

\section{System Prompts for Multi-Agent Pipeline}
\label{sec:agent_prompts}
In this section, we provide the complete system prompts for the three agents in our multi-agent pipeline. These prompts define the behavioral specifications that govern each agent's decision-making. By making these prompts explicit, we ensure full reproducibility and direct alignment with the main paper. Importantly, these prompts focus exclusively on structural and consistency-related instructions, avoiding aesthetic or cinematographic directives that might bias generation toward specific visual styles.

\begin{tcolorbox}[title=Storyboard Agent: Script-to-Shot Decomposition, breakable]
\textbf{Role} \\
You expand a script synopsis into an ordered list of shot descriptions.

\textbf{Goal} \\
Process the entire script synopsis ONCE and output a storyboard JSON
containing all shots. Each shot description must clearly state:
\begin{itemize}
    \item what happens,
    \item where it happens,
    \item who appears,
    \item which props appear,
\end{itemize}
so that downstream agents can process ONE shot at a time.

\textbf{Input} \\
- Script synopsis S (or a structured screenplay derived from S).

\textbf{Output} \\
A list (storyboard.json) of shot descriptions, each with:
\begin{itemize}
    \item scene
    \item scene\_description
    \item plot
    \item characters
    \item key\_props
    \item environment\_info (time jumps, time period)
\end{itemize}

\textbf{Core Rules}
\begin{enumerate}
    \item Do NOT invent new characters, props, locations, or events.
    \item Keep all descriptions factual and faithful to the original story.
    \item Make sure every shot lists all characters and props that appear on screen.
    \item Explicitly state any time jumps (e.g., older age).
    \item Do NOT include visual style, composition, or camera terms.
\end{enumerate}
\end{tcolorbox}

\begin{tcolorbox}[title=Memory Agent: Dynamic Memory Management, breakable]
\textbf{Role} \\
You are the Memory Agent. You maintain a Dynamic Memory Bank of characters, props, and scenes. For each shot, you decide whether to REUSE an existing reference image or CREATE a new one using the image generation tool.

\textbf{Image Generation Tool (for new references)} \\
You can request a new reference image by producing:
\begin{itemize}
    \item a short natural-language prompt describing the entity’s current state,
    \item optional reference image(s) from previous states of the same entity,
    \item a target image path.
\end{itemize}
The backend image generator ALWAYS returns an image file to that path.

\textbf{Goal} \\
For each single shot:
\begin{enumerate}
    \item Read the shot description.
    \item Extract all characters, props, and backgrounds that appear.
    \item Build a short text description of each entity’s current state.
    \item Retrieve or generate a reference image.
    \item Update the Dynamic Memory Bank.
\end{enumerate}

\textbf{Input} \\
Per call, you receive:
\begin{itemize}
    \item one shot description from \texttt{storyboard.json}
    \item the current Dynamic Memory Bank:
    \begin{itemize}
        \item character\_memory
        \item prop\_memory
        \item background\_memory
    \end{itemize}
\end{itemize}

\textbf{Core Steps (per shot)}
\begin{enumerate}
    \item \textbf{Entity extraction}
    \begin{itemize}
        \item Identify all entities that appear in this shot.
        \item Build a text state description for each entity (age, clothing, prop condition, time-of-day, etc.).
    \end{itemize}
    \item \textbf{Retrieval}
    \begin{itemize}
        \item In the appropriate memory bank, check if a semantically similar entity state already exists.
        \item If yes, REUSE its reference\_image\_path.
    \end{itemize}
    \item \textbf{Generation \& Update}
    \begin{itemize}
        \item If no suitable entry exists:
        \begin{itemize}
            \item Create a NEW reference image using the image generation tool.
            \item Use the current state description, plus any earlier images of this entity to preserve identity.
            \item Insert a new memory entry (name, state description, image path).
        \end{itemize}
    \end{itemize}
    \item \textbf{Temporal logic}
    \begin{itemize}
        \item If the story implies a time jump or meaningful change (older age, new period, changed prop condition, changed environment), generate a new entry.
        \item If no meaningful change, reuse the previous entry.
    \end{itemize}
\end{enumerate}

\textbf{Output Per Shot} \\
Return for THIS shot:
\begin{itemize}
    \item a list of entities with:
    \begin{itemize}
        \item entity\_name
        \item entity\_type (character, prop, background)
        \item state\_description
        \item reference\_image\_path
    \end{itemize}
    \item the updated Dynamic Memory Bank.
\end{itemize}
You always process exactly ONE shot per call.
\end{tcolorbox}

\begin{tcolorbox}[title=Visualization Agent: Keyframe \& Video Generation, breakable]
\textbf{Role} \\
You are the Visualization Agent. You generate:
\begin{enumerate}
    \item a keyframe image for this shot, and
    \item a short video clip based on that keyframe,
\end{enumerate}
using two distinct backend tools.

\textbf{Keyframe Generation Tool} \\
You can request a keyframe image by providing:
\begin{itemize}
    \item a natural-language prompt describing the combined scene,
    \item a list of reference images (characters, props, background) that must be respected,
    \item a target image path.
\end{itemize}
The backend will synthesize a still keyframe image at that path.

\textbf{I2V Video Generation Tool} \\
You can request a shot video by providing:
\begin{itemize}
    \item the keyframe image path (treated as the first frame),
    \item a short natural-language description of the shot’s action,
    \item a target video path.
\end{itemize}
The backend I2V model will generate a short video clip.

\textbf{Goal} \\
For EACH shot:
\begin{itemize}
    \item Use the shot description + reference images from the Memory Agent.
    \item Produce a keyframe prompt and image path.
    \item Produce a video prompt and save path.
    \item Keep characters, props, and scenes consistent with reference images.
\end{itemize}

\textbf{Input} \\
Per call, you receive:
\begin{itemize}
    \item one shot description from \texttt{storyboard.json}
    \item entity reference images (characters, props, background) for THIS shot
\end{itemize}

\textbf{Core Steps (per shot)}
\begin{enumerate}
    \item \textbf{Keyframe Construction}
    \begin{itemize}
        \item Combine the shot description and reference images into a single natural-language keyframe prompt.
        \item This prompt must respect the identity and appearance of every entity.
        \item Specify the output path for the keyframe image.
    \end{itemize}
    \item \textbf{Video Construction}
    \begin{itemize}
        \item Treat the keyframe as the FIRST frame.
        \item Write a concise prompt ($\le$ 500 characters) describing only:
        \begin{itemize}
            \item the essential action in this shot,
            \item how the moment evolves over time.
        \end{itemize}
        \item Specify the output path for the shot video.
    \end{itemize}
\end{enumerate}

\textbf{Core Rules}
\begin{enumerate}
    \item Process exactly ONE shot per call.
    \item Do NOT add new characters, props, or locations.
    \item Do NOT change identity, appearance, or layout from the reference images.
    \item Prompts must be short and factual; do NOT include style, composition, or camera specification.
    \item The video must remain consistent with the keyframe.
\end{enumerate}

\textbf{Output Per Shot} \\
A structured record with:
\begin{itemize}
    \item keyframe\_generation\_prompt
    \item keyframe\_image\_path
    \item video\_generation\_prompt
    \item video\_save\_path
    \item optional negative\_prompt
\end{itemize}
\end{tcolorbox}

\section{Prompt for Benchmark Story Generation}

\label{sec:benchmark_prompt}

In this section, we present the detailed prompt used to generate the 54 benchmark scripts. This prompt enforces specific constraints on story structure, shot count, and element persistence to ensure a rigorous evaluation of cross-shot consistency across varying narrative complexities.

\begin{tcolorbox}[title=Prompt for Benchmark Story Generation, breakable]
\textbf{Role} \\
You are a \textbf{professional film screenwriter}.

\textbf{Task} \\
Generate 54 short story scripts (6 separate samples for each of the 9 combinations specified below).

\textbf{Combinations} \\
We consider the following combinations of total shot count and core element persistence:
\begin{itemize}
    \item (4 Shots, Character Persistent)
    \item (4 Shots, Prop Persistent)
    \item (4 Shots, Scene Persistent)
    \item (8 Shots, Character Persistent)
    \item (8 Shots, Prop Persistent)
    \item (8 Shots, Scene Persistent)
    \item (12 Shots, Character Persistent)
    \item (12 Shots, Prop Persistent)
    \item (12 Shots, Scene Persistent)
\end{itemize}

\textbf{Input (Constraints)}
\begin{itemize}
    \item \textbf{Structure and Length}
    \begin{itemize}
        \item \textbf{Total Shots:} Each story must have exactly 4, 8, or 12 shots.
        \item \textbf{Three-Act Structure:} The story plot must be clearly distributed across all shots and explicitly divided into three acts. The number of shots per act must be allocated rationally based on the chosen total (e.g., Act 1: 20--30\%, Act 2: 40--60\%, Act 3: 20--30\%).
    \end{itemize}

    \item \textbf{Character and Prop}
    \begin{itemize}
        \item \textbf{Character:} There must be only one character, whose age must be greater than 20.
        \item \textbf{Prop:} There must be only one prop.
        \item \textbf{Prop Appearance and Size:} The prop's color must be a bright color (e.g., bright red, vivid yellow, neon green) with high contrast. The prop cannot be too small (e.g., at least as large as a backpack or a large book), must be visually prominent, and must be easy to separate from the background.
    \end{itemize}

    \item \textbf{Content Tone}
    \begin{itemize}
        \item \textbf{Family-Friendly:} The story must be family-friendly.
        \item \textbf{Forbidden Content:} All violent, gory, sexual, or disturbing content is strictly forbidden.
    \end{itemize}
\end{itemize}

\textbf{Output (Format Requirements)}
\begin{itemize}
    \item \textbf{Description Style (Critical)}
    \begin{itemize}
        \item \textbf{Natural Description:} Shot descriptions must be natural screenplay scene descriptions. They should focus on what is seen on screen, what the character does, or what happens.
        \item \textbf{Strictly Forbidden:} The use of technical camera terms is strictly forbidden (e.g., ``close-up'', ``pan'', ``camera follows'', etc.).
    \end{itemize}

    \item \textbf{Formatting}
    \begin{itemize}
        \item \textbf{Output Shot Content Only:} Only output the content for each shot.
        \item \textbf{Merge as Paragraphs:} Each shot must be a single, separate paragraph; bullet points must not be used.
        \item \textbf{Shot Numbering:} Each paragraph must begin with \texttt{SHOT [Number]} (e.g., \texttt{SHOT 1}, \texttt{SHOT 2}, \dots).
    \end{itemize}
\end{itemize}

\textbf{Notes (Critical Constraints)}
\begin{itemize}
    \item \textbf{Core Element Persistence}\\
    Each script must strictly follow one of the following three conditions. When one element (character, prop, or scene) remains consistent in all shots, the other two elements must change across shots.
    \begin{itemize}
        \item \textbf{Character Persistent:} The character must appear in all shots, and the character's appearance must remain unchanged.
        \begin{itemize}
            \item \textbf{Scene:} The scene must be different in each shot.
            \item \textbf{Prop:} The prop must change status across shots (e.g., appear in some shots and not in others).
        \end{itemize}

        \item \textbf{Prop Persistent:} The prop must appear in all shots, and the prop's appearance must remain unchanged.
        \begin{itemize}
            \item \textbf{Scene:} The scene must be different in each shot.
            \item \textbf{Character:} The character must change status across shots (e.g., appear in some shots and not in others).
        \end{itemize}

        \item \textbf{Scene Persistent:} All shots must take place in the same scene (location).
        \begin{itemize}
            \item \textbf{Character:} The character must change status across shots (e.g., appear or disappear).
            \item \textbf{Prop:} The prop must change status across shots (e.g., appear or disappear).
            \item \textbf{Example Pattern:} The status changes for character and prop must provide variation, for example: \texttt{SHOT 1 -- Char+Prop, SHOT 2 -- Char only, SHOT 3 -- Prop only, SHOT 4 -- Neither}.
        \end{itemize}
    \end{itemize}

    \item \textbf{Formatting Requirements (Critical)}
    \begin{itemize}
        \item \textbf{First Appearance:} When the character or prop first appears, a detailed description must be provided in parentheses right after its name.
        \begin{itemize}
            \item \textbf{Character Description Format:} (Age [must be $>$ 20] + appearance and clothing description).
            \item \textbf{Prop Description Format:} (Detailed appearance description, including bright color and relative size).
        \end{itemize}

        \item \textbf{Subsequent Appearances (Critical):} After the first appearance, the character's and prop's appearance (including clothing) must remain unchanged (e.g., the character does not get dirty or change clothes; the prop does not break). Therefore, it is strictly forbidden to use parentheses to describe the character or prop again in subsequent shots.
    \end{itemize}
\end{itemize}
\end{tcolorbox}

\begin{figure*}[t]  
    \centering
    \includegraphics[width=0.8\linewidth]{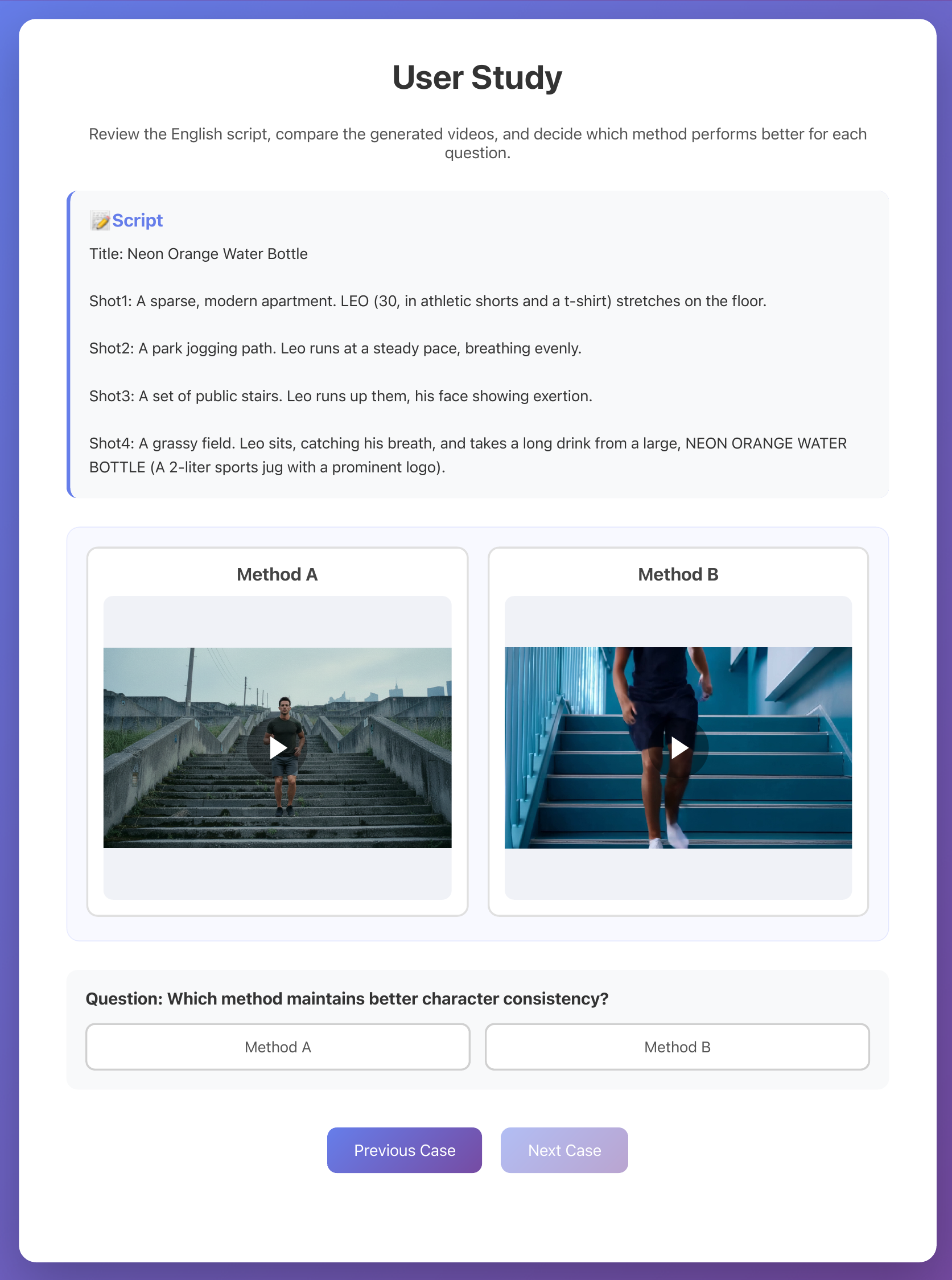}
    \caption{\textbf{Interface of the perceptual user study.}
    For each trial, participants first read the English script for a benchmark
    case (top), including the title and shot-level descriptions. They then view
    two generated videos corresponding to the same script, anonymized as
    \emph{Method A} and \emph{Method B} and shown side-by-side. At the bottom,
    a subclass-specific question is displayed (here: ``Which method maintains
    better character consistency?''). Participants answer via a forced choice
    by clicking one of the two method buttons.}
    \label{fig:user_study_interface}
\end{figure*}

\section{User Study}
\label{sec:supp_user_study}

In this section we provide additional details about the perceptual user study
summarized in the main paper. The goal of the study was to verify that the
improvements observed in our quantitative metrics are also perceived by human
observers.

\paragraph{Participants.}
We recruited 5 expert raters who routinely work with generative models and
video editing. All participants were fluent in English and were familiar with
basic cinematography concepts. They were instructed that the task focused
purely on evaluating \emph{consistency across shots}, rather than overall
aesthetics or story quality.

\paragraph{Stimuli.}
For each of the 54 benchmark scripts, we generated videos using our method and
five baseline methods, resulting in 270 method pairs (1--to--1 comparisons of
our method against each baseline for every script). Every video pair in the
study always corresponded to the same underlying script, shot breakdown, and
text prompts; only the generation method differed. The videos were shown in
their original resolution and aspect ratio and were trimmed to the same
duration used in our quantitative benchmark.

\paragraph{Interface.}
Figure~\ref{fig:user_study_interface} shows the web interface used in the
experiment. At the top of the page, participants saw the full English script
for the current benchmark case, including the title and the shot-level
descriptions (Shot 1--4/8/12). Below the script, two videos were displayed
side-by-side, labeled only as \emph{Method A} and \emph{Method B} to anonymize
the underlying algorithms. Participants could play, pause, and replay each
video as many times as they wished before responding. At the bottom of the
screen, a question was shown together with two large buttons labeled
``Method~A'' and ``Method~B'' for the forced-choice response. Navigation
buttons (``Previous Case'' and ``Next Case'') allowed participants to move
through the trials at their own pace.

\paragraph{Task and questions.}
Each trial consisted of a single method pair and a single
\textit{subclass-specific} consistency question. For every trial, participants
were asked one of the following three prompts:
\begin{itemize}
    \item \textbf{Character consistency:} ``Which method maintains better character
    consistency across all shots?''
    \item \textbf{Prop consistency:} ``Which method maintains better prop
    consistency across all shots?''
    \item \textbf{Background consistency:} ``Which method maintains better
    background consistency across all shots?''
\end{itemize}
Participants were explicitly told that there was no ``tie'' option and that
they should choose the video that better matched the question, even if the
difference was subtle. The question type for each trial was sampled uniformly
from the three subclasses, so that all methods were evaluated under all
consistency notions across the full set of benchmark cases.

\paragraph{Design and procedure.}
Each of the 5 participants evaluated 270 video pairs in total. The order of
trials was randomized independently for every participant. For each trial, the
left/right assignment of ``Method~A'' and ``Method~B'' was randomized to avoid
side bias. Participants completed the study remotely using their own displays
and were allowed to take breaks at any time; there was no time limit per
trial. Before starting the full study, each participant completed a short
practice block with example pairs not used in the benchmark to familiarize
themselves with the interface and question types.

\paragraph{Aggregation.}
For each benchmark script, baseline, and consistency subclass, we computed the
preference ratio as the fraction of trials in which our method was selected
over the baseline. These aggregated preference ratios are reported in the main paper. Consistent with the quantitative
metrics, our approach was strongly preferred across all subclasses, with
preference rates reaching up to 95.8\% for character and background
consistency. This confirms that the improvements measured by our automatic
metrics translate into a clearly perceptible increase in cross-shot coherence
for human viewers.


\section{Successful Cases}

\begin{figure*}[t]
    \centering
    \includegraphics[width=0.9\linewidth]{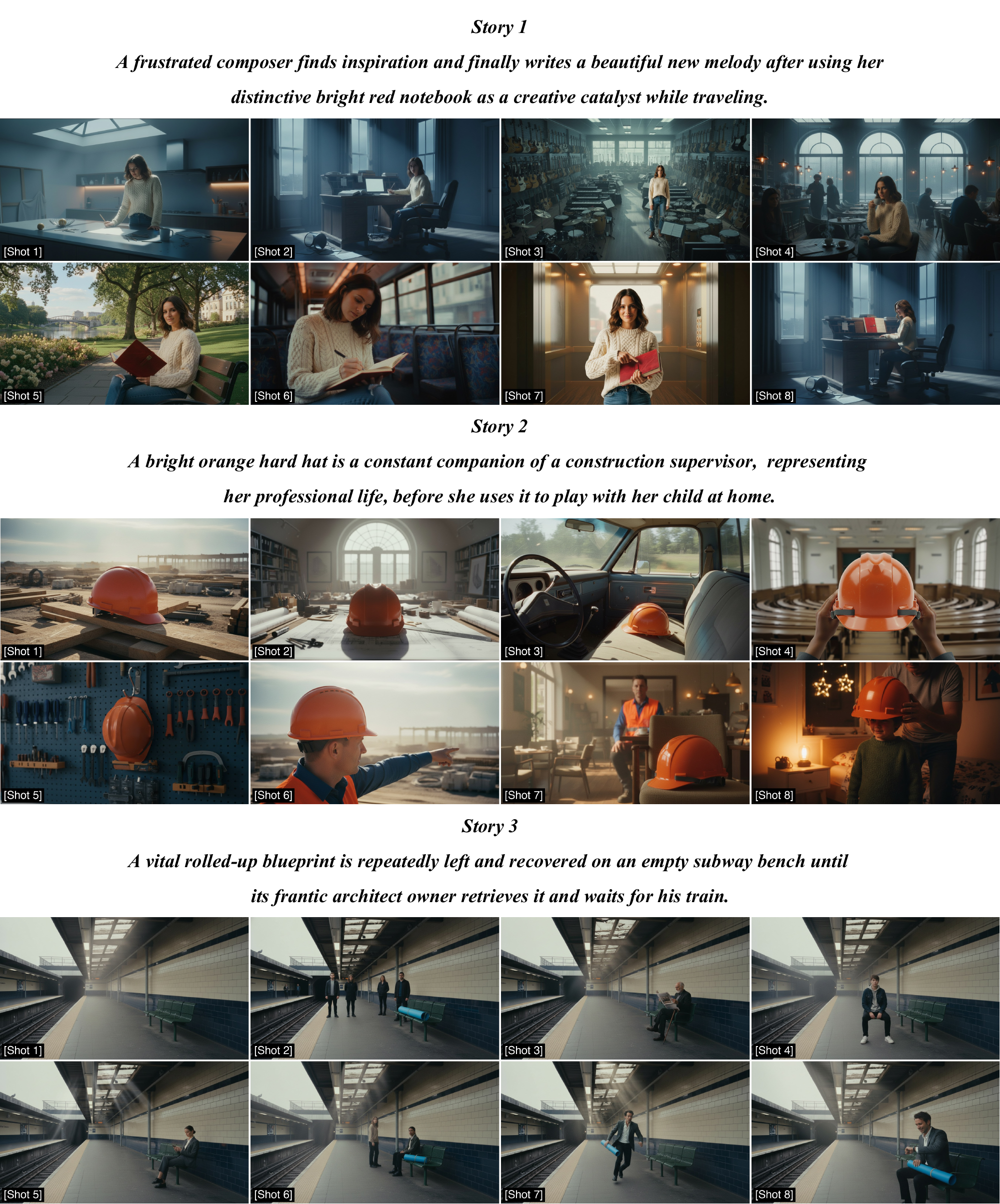}
    \caption{\textbf{Additional qualitative examples.} We visualize three generated stories from our benchmark, showcasing the three different modes of consistency. \textbf{Top (Story 1):} Character Consistency, where the main character is preserved across changing environments. \textbf{Middle (Story 2):} Prop Consistency, where the appearance of a specific object (hard hat) is maintained. \textbf{Bottom (Story 3):} Background Consistency, where the scene (subway station) remains fixed.}
    \label{fig:successful_case}
    \vspace{-0.2em}
\end{figure*}

Figure~\ref{fig:successful_case} provides additional qualitative results demonstrating our method's versatility across three consistency modes. Story 1 (top rows) showcases Character Consistency, maintaining the protagonist's identity across diverse settings including indoor rooms, outdoor parks, and public transit. Story 2 (middle rows) illustrates Prop Consistency, where a bright orange hard hat preserves its color and structure despite changing lighting and camera angles. Story 3 (bottom rows) demonstrates Background Consistency, with a subway platform remaining stable as a fixed backdrop for multiple characters and actions.
\clearpage

\clearpage

{
    \small
    \bibliographystyle{ieeenat_fullname}
    \bibliography{main}
}
\end{document}